# Semi-DerainGAN: A New Semi-supervised Single Image Deraining Network

Yanyan Wei[1], Zhao Zhang[1], Yang Wang[1], Haijun Zhang[2], Mingbo Zhao[3], Mingliang Xu[4], and Meng Wang[1]

[1] School of Computer Science and Information Engineering, Hefei University of Technology
[2] Department of Computer Science, Harbin Institute of Technology (Shenzhen)
[3] Donghua University & City University of Hong Kong
[4] School of Information Engineering, Zhengzhou University, China
E-mail: cszzhang@gmail.com

## ABSTRACT

Removing the rain streaks from single image is still a challenging task, especially for the cases that paired data are limited, since the shapes and directions of rain streaks in the synthetic datasets are very different from those of real images. Although supervised deep deraining networks have obtained impressive results using paired data of synthetic datasets, they still cannot obtain satisfactory results on real rainy images for the weak generalization of rain removal capacity, i.e., the pre-trained models usually cannot handle new shapes and directions that may lead to over-derained/under-derained results. In this paper, we mainly discuss the partial paired data-based SID and propose a new semi-supervised GAN-based deraining network called Semi-DerainGAN, which can use both synthetic and real rainy images in a uniform network based on two supervised and unsupervised processes. For this task, a semi-supervised rain streak learner termed SSRML sharing the same parameters of both processes is derived, which makes the real images contribute more rain streak information, so that the resulted model has a strong generalization power to the real SID task. To obtain better deraining results, we design a paired discriminator for distinguishing the real pairs from fake pairs. We also contribute a new real-world rainy image dataset called Real200 to alleviate the difference between both synthetic and real image domains. Extensive results on public datasets show that our model can obtain competitive or even better performance, especially on real images.

## KEYWORDS

Semi-supervised GAN-based SID network; semi-supervised rain streak learner; deep representation learning; image restoration

## 1 INTRODUCTION

Singe image deraining (SID), which is as a classical image restoration task, has been a challenging and interesting topic in the areas of computer vision and artificial intelligence due to its broad realistic applications, such as drone-based video surveillance, real-time object recognition and autonomous cars. SID mainly discussed the issue of modeling the rain streak information and recovering the background. The problem of SID can be formulated as follows:

$$X = R + B, \qquad (1)$$

where $X$ denotes a rainy image that will be decomposed into a rain-streak component $R$ and a clean background $B$, i.e., removing the rain streaks from the rainy images. Because Eqn. (1) is an ill-posed problem, some feasible approaches have been proposed to solve it, include both the traditional and deep learning-based models, which will be introduced briefly in Section 2.

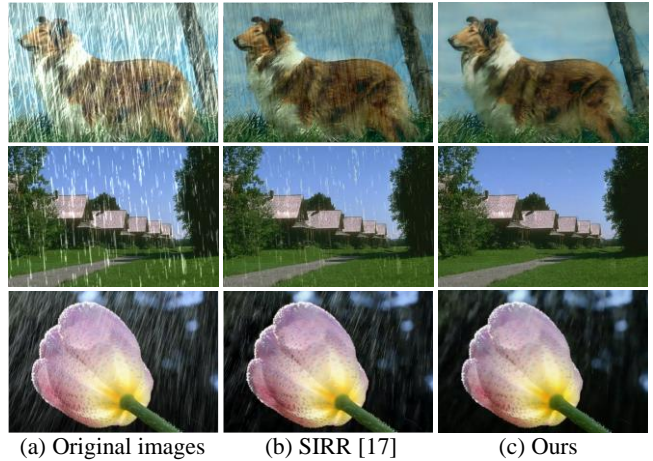

(a) Original images (b) SIRR [17] (c) Ours

**Figure 1:** Comparison of the semi-supervised deraining results of SIRR and our Semi-DerainGAN. From top to bottom are based on the synthetic images of Rain100H [19] and Rain100L [19], and the real images of SIRR [17]. We see that our model performs better, while SIRR leaves more rain streak information.

It is noteworthy that most existing deep deraining networks are supervised methods using paired information in synthetic datasets. The strong constraint can make the network convergence fast on the synthetic images, however, the performance on the real rainy images is still unsatisfactory due to lack of paired information. To address this issue, some researchers start to shift to study the semi-supervised deraining models, which can also use real images to enhance the model generalization ability. For example, a semi-supervised SID framework called Semi-Supervised Transfer Learning (or shortly SIRR) [17] was recently proposed to address the paired data restricted SID task. Different from the traditional supervised SID methods that are equipped with all paired data in the synthetic dataset, SIRR further adds the real rainy images without ground-truth for the network training. This is achieved by taking the residual between the input rainy image and expected network output (clear image without rain) as a specific parameterized rain fringe distribution. As such, SIRR adapts to the real unsupervised rain types by transferring the supervised synthetic rain, which can clearly alleviate the issue of insufficient training data and supervised sample bias. But the difference of the rain streaks in synthetic and real images is very large, so it may be inappropriate to train the model by adding a relatively hard constraint between both synthetic and real rainy image domains in a single network, i.e., minimizing the KL divergence between a synthetic rain streak mask and the real one learned from real images via training. Note that the KL divergence is the most important part of SIRR, which prompts the semi-supervised model to be effective. As a result, SIRR may leave rain streak information based on the synthetic and real images (see Fig.1). The difference between our model and SIRR is also illustrated in Fig.2.

In this paper, we investigate how to improve the paired data restricted SID task and propose a new semi-supervised SID network that use two processes to train synthetic and real images respectively via a hybrid loss. The main contributions are summarized as:

(1) A new and effective semi-supervised single image deraining network, termed Semi-DerainGAN, is technically proposed. Semi-DerainGAN can use partial paired data for SID, i.e., using both synthetic rainy images and real rainy images in a uniform network by two separate processes (i.e., supervised and unsupervised ones). Due to the designed two processes, Semi-DerainGAN can clearly avoid the interaction between synthetic and real images to be degraded, suffered by the single network of SIRR, which is cuased by the large difference between the rain streak information of the synthetic and real image domains. To the best of our knowledge, this is one of few semi-supervised deep models for SID.

(2) To encode the rain steak information more accurately, Semi-DerainGAN designs the semi-supervised model using two supervised and unsupervised processes. Specifically, the supervised process is used for the synthetic image training with paired data, while the unsupervised process is designed for real image training using unpaired data. More importantly, a new semi-supervised rain mask learner (*SSRML*) is designed to learn rain streak information from both synthetic and real rainy images, and two generators for two independent processes are used to generate the derained images by training. Note that the designed semi-supervised process of Semi-DerainGAN can clearly alleviate the degradation issue caused by the difference between synthetic and real rain streak information.

(3) A new real rainy image dataset Real200 is also created, which can be applied by other models to alleviate the difference between synthetic and real image domains. To the best of our knowledge, this is the first real rainy image dataset with more complex carefully-designed rain steak directions and shapes. Due to the attractive properties of Real200, we find experimentally that it can improve the semi-supervised SID task effectively (see Table 4).

(4) Extensive experiments on several challenging synthetic datasets and real images demonstrate the effectiveness of our Semi-DerainGAN for semi-supervised SID, compared with the most related semi-supervised method SIRR.

## 2 RELATED WORK

To solve the SID problem, many traditional algorithms have been proposed, such as low-rank representation methods [20], Gaussian mixture models [10], and sparse coding-based models [11]. In recent years, some deep learning-based deep network models have been proposed. For example, a contextualized dilated network [19] was recently proposed to jointly detect and remove the rain streaks from single image. The authors of [2] use the residual block [5] to reduce the mapping range from the input to output directly, which makes the learning process easier. A novel density-aware multi-stream dense convolutional network-based framework [21] was proposed to jointly estimate the rain density and deraining. More recently, a hybrid block [18] has been proposed to extract the rain streak more precisely, especially in heavy rainy condition. The authors of [32] propose a novel residual-guide feature fusion network which uses the recursive convolution to generate from shallower blocks to guide deeper blocks. Similarly, a better and simpler baseline deraining network is proposed in [24] by repeatedly unfolding a shallow ResNet to take advantage of the recursive computation. A dual-task network is also proposed in [31], which has two sub-networks: GAN and CNN, to remove the rain streaks by coordinating two mutually exclusive objectives self-adaptively. Also, there are some other deraining networks proposed recently [25-30].

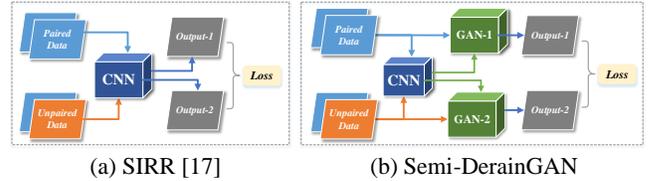

(a) SIRR [17]  (b) Semi-DerainGAN

**Figure 2:** Illustration of the semi-supervised network structures of (a) SIRR and (b) Semi-DerainGAN for comparison, from which we see that our network has a shared CNN and two independent GAN for paired and unpaired data, while SIRR only applies a single CNN and hence causing the information loss when training on paired data.

Because of the powerful capacity of generating realistic images, Generative Adversarial Networks (GAN) [3] have achieved superior performance in many vision tasks, including the SID. [22] proposed a conditional GAN-based network model which considers quantitative, visual and discriminative performance into the objective function. [13] proposed a raindrop removal method which uses GAN to produce the attention map by an attentive-recurrent network and uses it to generate a raindrop-free image through a contextual auto-encoder. Note that these existing GAN-based models are supervised methods that need paired data for training, so they are hard to be used in both the semi-supervised and unsupervised modes. To solve the restrictions on the conditions of using GAN, CycleGAN [23] was proposed by using unpaired data in training process, which can translate an image from a source domain $X$ to a target domain $Y$ and then reconstruct it from the domain $Y$ to domain $X$, formulated as $X \rightarrow Y \rightarrow X$. In our network, we use the CycleGAN for the unsupervised learning process.

## 3 PROPOSED METHOD

### 3.1 Network Architecture

We show the framework of Semi-DerainGAN in Fig.3, which has two supervised and unsupervised processes. More specifically, the framework has a semi-supervised rain mask learner (*SSRML*), three generators ($G_s, G_r, G_r^{'}$) that dispose on synthetic rainy images and real rainy images respectively, and three discriminators ($D_s, D_r, D_p$) corresponding to the generators. The synthetic and real rainy images are denoted as $\{x_s^i, y_s^i\}_{i=1}^{M_s} \in S$ and $\{x_r^i, \hat{y}_r^i\}_{i=1}^{N_r} \in R$, respectively, where $x_s$ and $y_s$ are rainy image and its corresponding rain-free image (i.e., label) in synthetic dataset $S$, while $x_r$ and $\hat{y}_r$ are rainy image and corresponding rain-free image in real dataset $R$. Since $x_r$ has no corresponding ground truth, we randomly choose $\hat{y}_r$ from

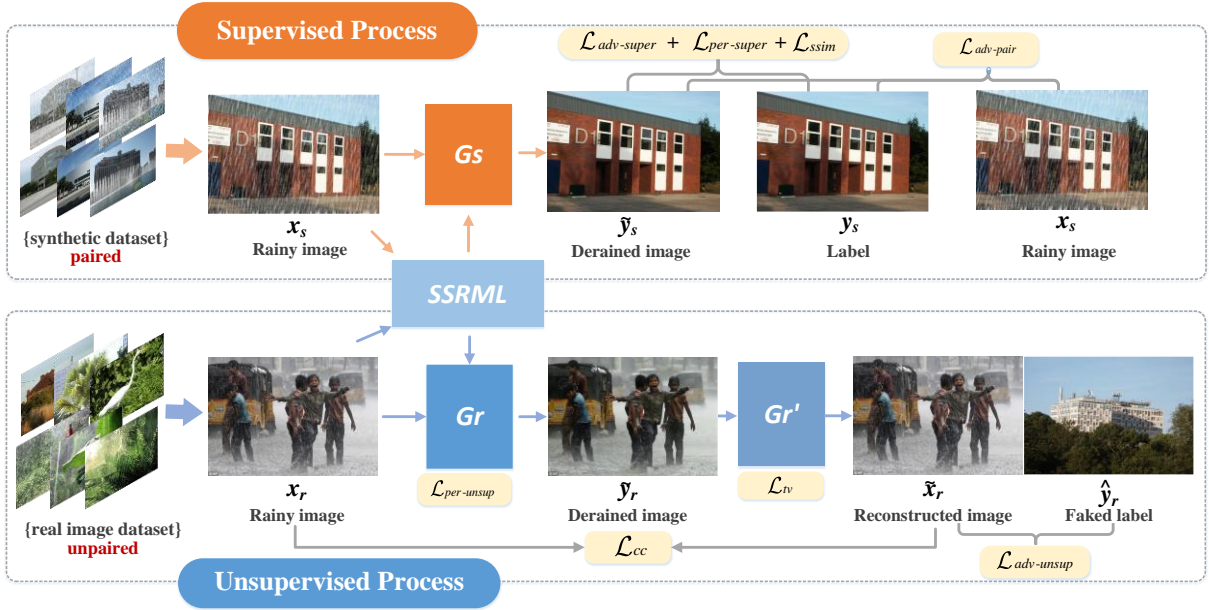

**Figure 3:** The framework of our Semi-DerainGAN that includes two processes, i.e., supervised and unsupervised processes. The supervised process is for synthetic image training by minimizing the losses of $\mathcal{L}_{super}$ denoted in orange color, while the unsupervised process is for real-world image training by minimizing the losses of $\mathcal{L}_{unsup}$ denoted in blue color. The semi-supervised rain mask learner SSRML is a shared module that can learn the rain streak information from both the synthetic and real-world rainy images.

$\{y_s^i\}_{i=1}^{M_s} \in S$ as the ground truth (i.e., fake label). In what follows, we will introduce detailed information of each part respectively.

**(1) Semi-Supervised Rain Mask Learner (SSRML)**

The original rain mask learner (*RML*), as an attentional rain drops extractor by [13], performs in supervised mode, which includes a *LSTM* unit and five *Conv-Relu-Conv-Relu* units. Due to the supervised nature of *RML*, it needs all paired data that are not easy to be obtained in reality, which will directly restrict its real applications. To address this issue, we design a structure to use *RML* as an attentional rain streak extractor in semi-supervised mode, called semi-supervised rain mask learner (*SSRML*), to learn rain streak information (i.e., shapes and directions) from both the synthetic and real-world rain image domains. Under the circumstances, *SSRML* can obtain better deraining result in the real-world SID task. Overall, this process can be formulated as

$$m_s = SSRML(x_s), \; m_r = SSRML(x_r), \quad (2)$$

where $m_s$ and $m_r$ are the rain masks extracted from synthetic rainy image $x_s$ and real rainy image $x_r$, respectively.

**(2) Generators in Synthetic and Real Image Domains**

We describe the three generators $G_s, G_r, G_r'$. $G_s$ and $G_r$ can generate the derained images from synthetic data $S$ and real data $R$ in training process respectively, and we utilize the U-net. Then, $G_r'$ reconstructs the real rainy image for a consist purpose which is proposed in CycleGAN [23], and is consisted of a *SSRML* and a U-net. The formulation for the generators can be represented as

$$\tilde{y}_s = G_s(m_s, x_s), \; \tilde{y}_r = G_r(m_r, x_r), \; \tilde{x}_r = G_r'(\tilde{y}_r), \quad (3)$$

where $\tilde{y}_s$ and $\tilde{y}_r$ are the derained results of images $x_s$ and $x_r$, and $\tilde{x}_r$ is reconstructed rainy image of the derained image $\tilde{y}_r$.

**(3) Discriminators in Synthetic and Real Domains**

In this section, we introduce the three discriminators $D_s, D_r, D_p$ in our model, which have two types of discriminators. The first type contains $D_s$ and $D_r$ that apply multi-scale structures, where the feature maps at each scale go through five convolution layers and then are fed into sigmoid outputs [9]. We use 3 different scales for $D_s$ and $D_r$. The second type of discriminator $D_p$ is a paired discriminator [4] proposed for the makeup transforming. In our network, we simplify the input of $D_p$ and use paired images, i.e., rainy and derained images, to make the network generate more realistic derained images. The efficiency of the paired discriminator is verified by simulations. The adversarial losses for used generators and discriminators can be defined as follows:

$$\begin{aligned}
\mathcal{L}_{adv\text{-}super} &= \mathbb{E}_{x_s \sim S}\left[\log D_s(y_s)\right] \\
&+ \mathbb{E}_{x_s \sim S}\left[\log\left(1 - D_s\left(G_s(SSRML(x_s), x_s)\right)\right)\right] \\
&+ \mathcal{L}_{adv\text{-}pair}
\end{aligned} \quad (4)$$

$$\begin{aligned}
\mathcal{L}_{adv\text{-}pair} &= \mathbb{E}_{x_s \sim S}\left[\log D_p(x_s, y_s)\right] \\
&+ \mathbb{E}_{x_s \sim S}\left[\log\left(1 - D_p\left(x_s, (G_s(SSRML(x_s), x_s))\right)\right)\right]
\end{aligned}$$

$$\begin{aligned}
\mathcal{L}_{adv\text{-}unsup} &= \mathbb{E}_{x_r \sim R}\left[\log D_r(\hat{y}_r)\right] + \\
&+ \mathbb{E}_{x_r \sim R}\left[\log\left(1 - D_r\left(G_r(SSRML(x_r), x_r)\right)\right)\right]
\end{aligned} \quad (5)$$

where $\mathcal{L}_{adv\text{-}super}$ is the adversarial loss which contains a loss of pair discriminator $\mathcal{L}_{adv\text{-}pair}$ for supervised process and $\mathcal{L}_{adv\text{-}unsup}$ is an adversarial loss for the unsupervised process.

**3.2 Supervised Process**

In the supervised process, we use the synthetic data $\{x_s^i, y_s^i\}_{i=1}^{M_s} \in S$ to learn the parameters of network, i.e., $SSRML, G_s, D_s$ and $D_p$. We minimize the following supervised loss function:

$$\mathcal{L}_{super} = \lambda_{adv\text{-}super}\mathcal{L}_{adv\text{-}super} + \lambda_{per\text{-}super}\mathcal{L}_{per\text{-}super} + \lambda_{ssim}\mathcal{L}_{ssim}, \quad (6)$$

where $\mathcal{L}_{adv\text{-}super}$ is an adversarial loss for the process in Eqn. (4), $\mathcal{L}_{per\text{-}super}$ denotes the perceptual loss [7] that can encode the difference between the derained image $\tilde{y}_s$ and the corresponding label image $y_s$, $\mathcal{L}_{ssim}$ is the SIMM loss [16] that can keep structural similarity between the two images, $\lambda_{adv\text{-}super}$, $\lambda_{per\text{-}super}$ and $\lambda_{ssim}$ are tradeoff parameters to control the contributions of each loss. Note that the two losses $\mathcal{L}_{per\text{-}super}$ and $\mathcal{L}_{ssim}$ can be defined as

$$\mathcal{L}_{per\text{-}super} = \|\partial_v(y_s) - \partial_v(\tilde{y}_s)\|_2^2, \quad (7)$$
$$\mathcal{L}_{ssim} = -SSIM(y_s, \tilde{y}_s), \quad (8)$$

where $\partial_v(\cdot)$ denotes the feature extractor of the $conv_{2,3}$ layer of the VGG-16 network [14] that are pre-trained on the ImageNet [1]. $SSIM(\cdot)$ is the SSIM function to calculate the similarity between two images $y_s$ and $\tilde{y}_s$. Note that we aim at maximizing the SSIM value as much as possible, so $\mathcal{L}_{ssim}$ is a negative function.

### 3.3 Unsupervised Process

In the unsupervised process, we use the real-world rain image data $\{x_r^i, \hat{y}_r^i\}_{i=1}^{N_r} \in R$ without paired information to learn the parameters of the network, i.e., $SSRML, G_r, G_r'$ and $D_r$. Specifically, we minimize the following unsupervised loss function:

$$\mathcal{L}_{unsup} = \lambda_{adv\text{-}unsup}\mathcal{L}_{adv\text{-}unsup} + \lambda_{cc}\mathcal{L}_{cc} + \lambda_{per\text{-}unsup}\mathcal{L}_{per\text{-}unsup} + \lambda_{tv}\mathcal{L}_{tv}, \quad (9)$$

where $\mathcal{L}_{adv\text{-}unsup}$ is the adversarial loss for unsupervised process, $\mathcal{L}_{cc}$ ensures that the derained image $\tilde{y}_r$ can be translated back to the original rainy image $x_r$ for preserving the contents of images, $\mathcal{L}_{per\text{-}unsup}$ similarly defined as $\mathcal{L}_{per\text{-}super}$, $TV(\cdot)$ is the TV function to make generated real derained image more realistic. $\lambda_{adv\text{-}unsup}$, $\lambda_{cc}$, $\lambda_{per\text{-}unsup}$ and $\lambda_{tv}$ are the trade-off parameters to balance the losses. Technically, $\mathcal{L}_{cc}$, $\mathcal{L}_{per\text{-}unsup}$ and $\mathcal{L}_{tv}$ can be defined as

$$\mathcal{L}_{cc} = \mathbb{E}_{x_r \sim R}[\|x_r - \tilde{x}_r\|_1], \quad (10)$$
$$\mathcal{L}_{per\text{-}unsup} = \|\partial_v(x_r) - \partial_v(\tilde{y}_r)\|_2^2, \quad (11)$$
$$\mathcal{L}_{tv} = TV(\tilde{y}_r). \quad (12)$$

### 3.4 Objective Function

The overall loss function of Semi-DerainGAN used for training the proposed network is defined as follows:

$$\mathcal{L}_{total} = \mathcal{L}_{super} + \lambda_{unsup}\mathcal{L}_{unsup}, \quad (13)$$

where $\lambda_{unsup}$ is a tradeoff parameter to balance the supervised process $\mathcal{L}_{super}$ and unsupervised process $\mathcal{L}_{unsup}$.

### 3.5 Created Real Rainy Image Dataset Real200

For the SID task, most existing rainy image datasets are synthetic, which contains less shape and density information about the rain streaks. Recently, the authors of [17] proposed a real rainy image dataset termed SIRR-Data, which has 147 real rainy images with more shape and density information than synthetic ones. However, SIRR-Data still have several drawbacks: (1) they contain many unsuitable weather conditions, e.g., snow weather and mist weather, which can hardly prompt the convergence in the semi-supervised SID training process and even has a negative effect; (2) they contain some low-resolution images whose rain streak information are difficult to be detected; (3) the real images of SIRR-Data are raw and not carefully prepared, as they may contain rain-free regions, or too light to see the rain streaks in sky or rainwater ground, or contain the rain drops or rain spray. Maybe some researchers think that these situations can enhance the generalization ability of the deep network, but in fact it may be negative for training due to the interference. Although these images are realistic, they are unsuitable for semi-supervised mode. In other words, if we could solve these drawbacks, we can train a better semi-supervised SID model under the same setting. As such, we create a new real rainy image dataset Real200. The images are carefully selected from Internet and manually cropped to highlight the rainy regions (see Fig.4).

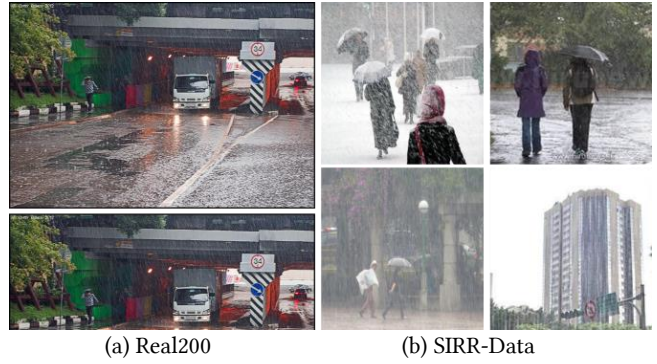

(a) Real200        (b) SIRR-Data

**Figure 4:** Real images in two datasets (SIRR-Data and Real200). In (a), the first row is the original real rainy image from Internet and the second row shows cropped ones suitable for semi-supervised learning. In (b), from the top left to bottom right are snow weather, low resolution, mist weather and bright region respectively.

The advantages of Real200 are twofold: 1) it only contains rainy conditions and all images are high-resolution; 2) it is cropped manually to highlight the rain streak regions, while preserving the diversity of the rain streaks in real images. Note that the rain streaks of Real200 are not significantly different from those in synthetic images, so it will be beneficial for training a semi-supervised model, as can be seen in Table 4. Due to the positive effect of Real200 dataset, it can be utilized in other semi-supervision SID tasks.

## 4 EXPERIMENTS AND ANALYSIS

### 4.1 Training Details

We describe the implementation details of our Semi-DerainGAN is trained by Pytorch [12] in Python environment on a NVIDIA GeForce GTX 1080i GPU with 12GB memory. We crop each rainy image to many patches of 100×100 by stride of 80. Adam [8] is used as the optimization algorithm with a batch size of 4. The model is trained for total 200 epochs. The learning rates of supervised and unsupervised processes are set to 1e-4 and 1e-3, respectively. The learning rate is decayed with a policy of Pytorch after 100 epochs. The trade-off parameters in Eqns. (6) and (13) are set to 1, and the trade-off parameters in Eqn. (9) are set to 1.5e-5, 10, 1 and 100, respectively. All the tradeoff parameters are chosen empirically.

**Table 1:** Comparison with each model in terms of PSNR on Rain1400 and Rain1400&SIRR-Data. **Bold** denotes the best performance. *w/o R* means that only using the synthetic dataset Rain1400, i.e., trained in supervised mode. Sparse and dense mean two different scenarios that have different complexity and multiformity of the rain streaks, which are detailed in SIRR [17]. Each scenario has 10 images.

| Dataset | Input | Rain1400 | | | | | Rain1400&SIRR-Data | | |
|---|---|---|---|---|---|---|---|---|---|
| | | *DSC* | *GMM* | *DDN* | *JORDER* | *DID-MDN* | *SIRR* | *Ours w/o R* | *Ours* |
| Sparse | 24.14 | 25.05 | 25.67 | 26.88 | 24.22 | 25.66 | 26.98 | 26.15 | **28.21** |
| Dense | 17.95 | 19.00 | 19.27 | 19.90 | 18.75 | 18.60 | 21.60 | 20.32 | **23.15** |

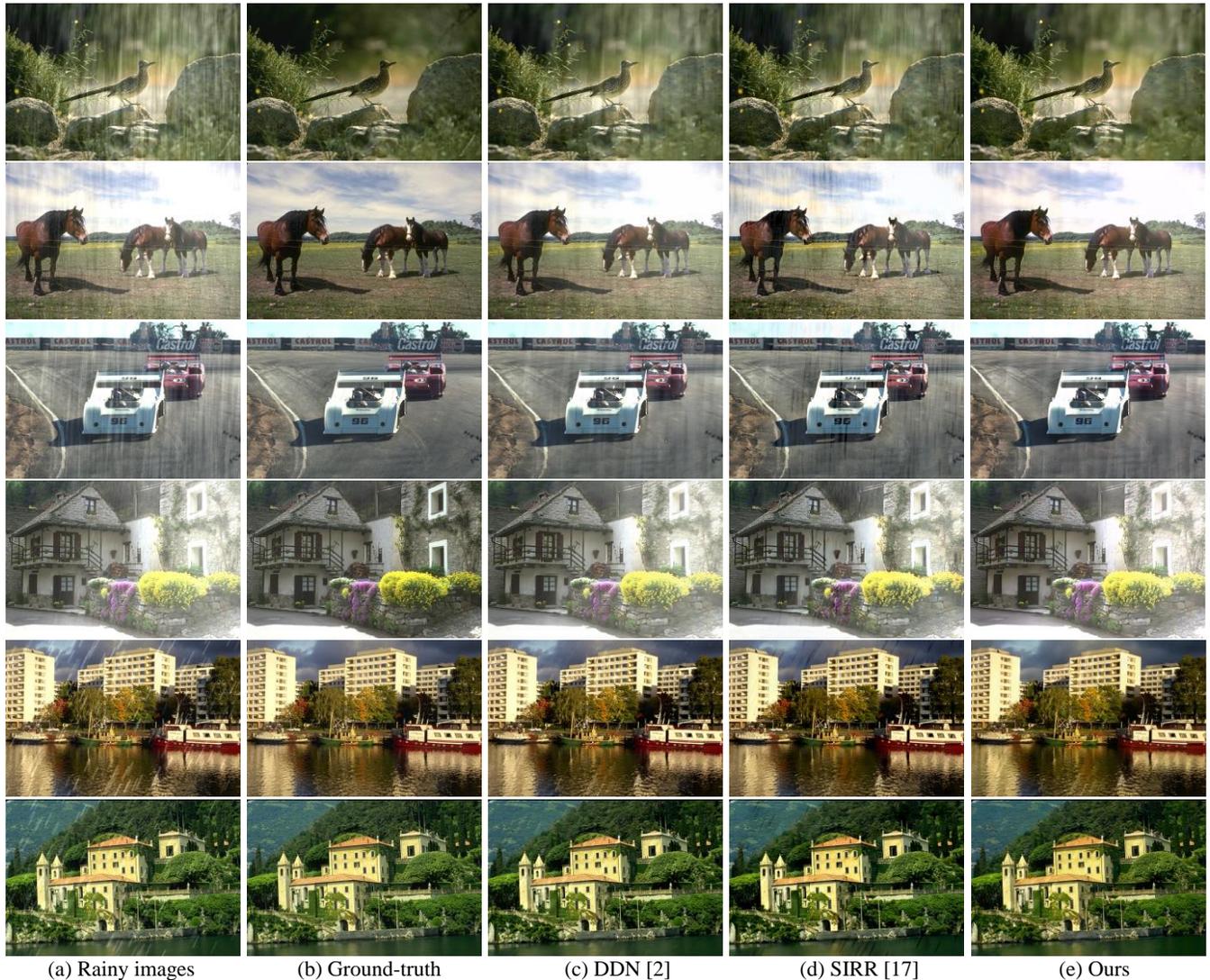

(a) Rainy images     (b) Ground-truth     (c) DDN [2]     (d) SIRR [17]     (e) Ours

**Figure 5:** Comparison of the deraining performance on the test images mentioned in SIRR [17], form which we see that our method obtains the best visual deraining performance, compared with several related deep models.

## 4.2 Datasets and Evaluation Metrics
**Evaluated Synthetic and Real Rainy image Datasets**

Four synthetic datasets are used: (1) Rain100H [19] that has five streak directions, contains 1,800 rainy images for training and 100 rainy images for testing; (2) Rain100L [19], where 200 image pairs are used for training and 100 image pairs are used for testing; (3) Rain1400 [2] has 14000 pairs of rainy images with 14 kinds of different rain streak orientations and magnitudes; (4) Rain12 [10] contains 12 rainy and clean image pairs. Since Rain12 has few samples, we directly use the trained network on Rain100L to test it.

For real images, we use two datasets: (1) SIRR-Data [17] that contains 147 real rainy images; (2) Real200 that is created in this paper, which contains 200 real-world rainy images. Since our method and SIRR are all semi-supervised methods, we use the synthetic dataset plus real-world dataset for training, which is denoted by &, such as Rain1400&SIRR-Data and Rain100L&Real200.

**Evaluation Metrics and Compared Methods**

The Peak Signal-to-Noise Ratio (PSNR) [6] and Structural Similarity Index (SSIM) [16] are used for evaluating the rainy images with ground-truth. For the real rainy images that have no ground-truth, such as SIRR-Data, we only provide the visual deraining results.

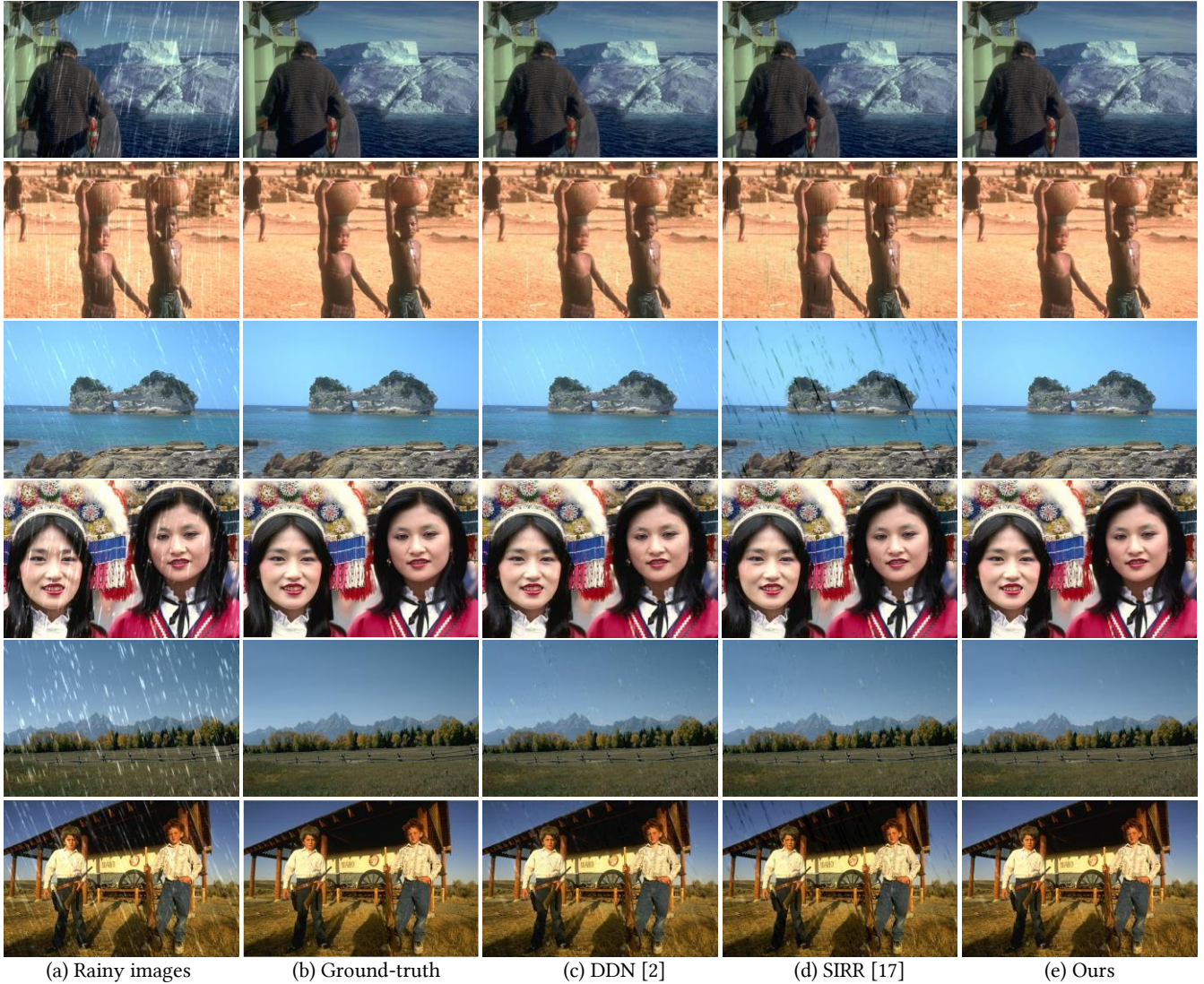

| (a) Rainy images | (b) Ground-truth | (c) DDN [2] | (d) SIRR [17] | (e) Ours |

**Figure 6:** Visual comparison of deraining results on Rain100L [19]. We see our method can wipe out most rain streak in rainy images.

**Table 2:** Quantitative comparison on the synthetic datasets. Since SIRR is a semi-supervised model, the comparison to it will be most meaningful. * denotes that the model trained on specified synthetic dataset plus SIRR-Data. **Bold** denotes the best performance.

| Da-tasets | Rain100H | | Rain100L | | Rain12 | |
|---|---|---|---|---|---|---|
| **Metrics** | PSNR | SSIM | PSNR | SSIM | PSNR | SSIM |
| *DSC* | 13.77 | 0.320 | 27.34 | 0.849 | 30.07 | 0.866 |
| *GMM* | 15.23 | 0.450 | 29.05 | 0.872 | 32.14 | 0.916 |
| *DDN* | 22.85 | 0.725 | 32.38 | 0.926 | 34.04 | 0.933 |
| *SIRR** | 22.47 | 0.716 | 32.37 | 0.925 | 34.02 | 0.935 |
| *Ours** | **22.89** | **0.801** | **34.12** | **0.958** | **35.86** | **0.960** |

In this experiment, six popular SID models are added for comparing with, including two model-driven methods (i.e., DSC [11] and GMM [10]), three supervised deep network models (i.e., DDN [2], JORDER [19] and DID-MDN [21], and one most related semi-supervised deep network model (i.e., SIRR [17]).

### 4.3 Results on Synthetic Rainy images

We first evaluate each method on the Rain1400&SIRR-Data, which contains a total 14000 pairs of synthetic rain images plus 147 real rainy images for the model training. Besides, we use other 20 images (10 sparse rain streak and 10 dense rain streak images), which are also employed by SIRR, for testing for fair comparison. The supervised models are directly trained on Rain1400, and the semi-supervised deep models are trained on Rain1400&SIRR-Data. For fair comparison, we directly use the deraining results of other methods from SIRR [17]. From Table 1, we see that our model can obtain the best performance on both sparse and dense test data. To show the effect of unsupervised process in our model, we also train it using a supervised mode totally (i.e., without real data). The result shows that the performance of adding real images into the training process can be improved. The illustration of the deraining results can also be seen in Fig. 5. We also evaluate several SID models on synthetic datasets (i.e., Rain100H, Rain100L and Rain12) in Table 2 and visualize some SID results in Fig.6. The image deraining results also demonstrate the superior performance of our proposed Semi-DerainGAN compared with other related methods.

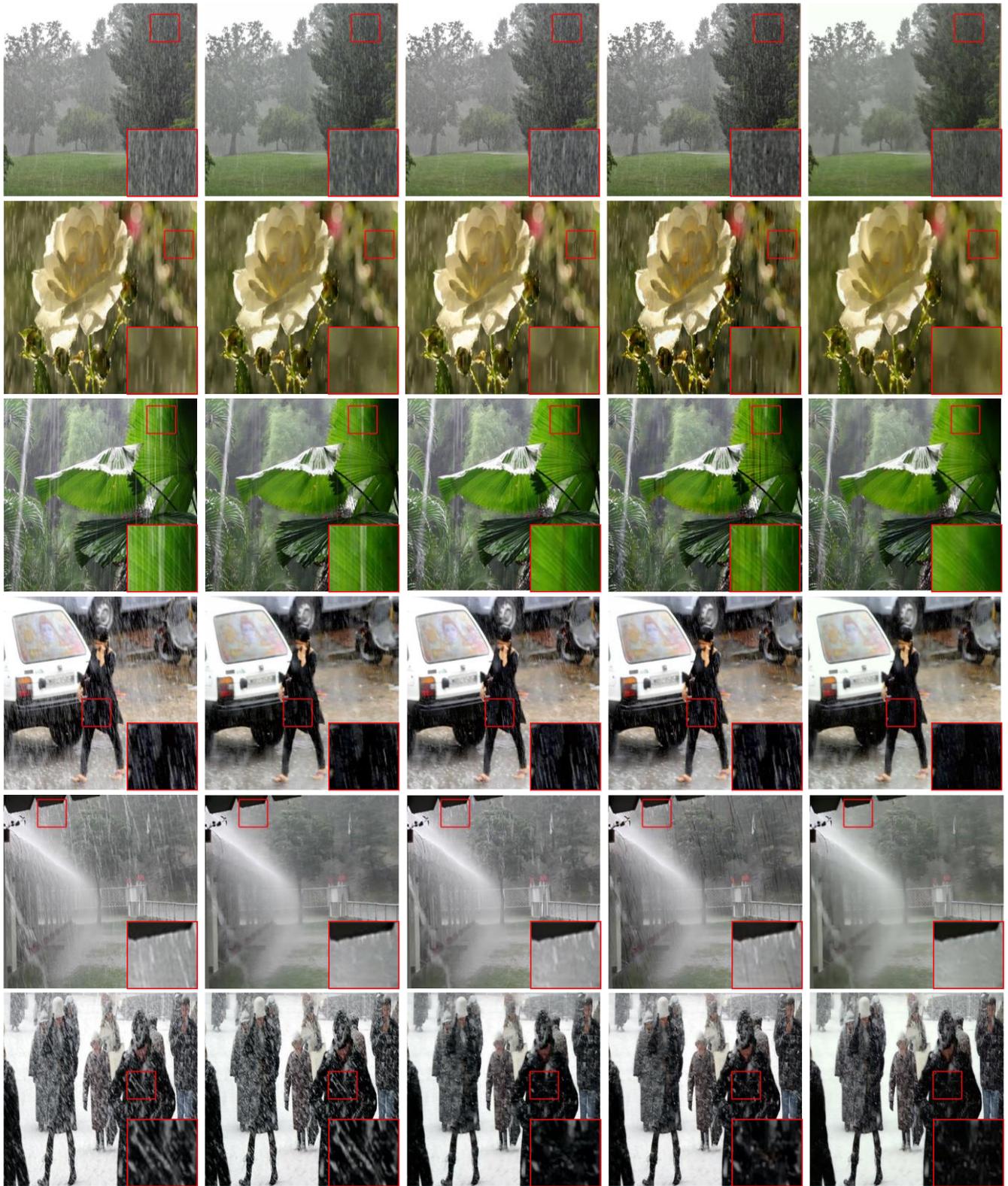

(a) Rainy images     (b) DDN [2]     (c) PReNet [24]     (d) SIRR [17]     (e) Ours

**Figure 7:** Visual deraining performance comparison of our Semi-DerainGAN with two supervised SID models (b) and (c), and one semi-supervised SID method (d), on the real-world rainy images.

**Table 3:** Deraining results of our Semi-DerainGAN with different settings on Rain100L&SIRR-Data, where w/o means without specified loss or module. **Bold** denotes the best performance.

| Set- | w/o. $\mathcal{L}_{per}$ | w/o. $\mathcal{L}_{tv}$ | w/o. $D_p$ | $\mathcal{L}_{total} + D_p$ |
|---|---|---|---|---|
| PSNR | 32.25 | 33.82 | 29.94 | **34.12** |
| SSIM | 0.939 | 0.944 | 0.949 | **0.958** |

### 4.4 Results on Real Rainy Images

We compare the semi-supervised deraining result of our network with those of DDN, PReNet, SIRR on Rain1400&SIRR-Data. We visualize the SID results in Fig.7, from which we see that our network performs better than the other methods, especially for the most related SIRR, since SIRR leaves more rain streak information, which keeps consistent with the above numerical evaluation results.

### 4.5 Ablation Study

We discuss the selections of loss functions and module in our network, and demonstrate the availability of our created real rain image dataset Real200. To reduce the time consumption for training, we simply use Rain100L& SIRR-Data as the training data.

**Loss Functions and Module**

We explore the deraining results of our Semi-DerainGAN without $\mathcal{L}_{per} = \mathcal{L}_{per\text{-}unsup} + \mathcal{L}_{per\text{-}super}$ or $\mathcal{L}_{tv}$ in Table 3, from which we see that the deraining performance goes down without partial loss, especially on the metric of PSNR. Besides, we also evaluate the importance of $D_p$ in our Semi-DerainGAN, and we can see that without $D_p$, the resulted performance drops fast. The results demonstrate that: 1) the design of the hybrid loss functions in our supervised and semi-supervised processes can optimize our network effectively. Since the perceptual loss constrains the feature maps between both paired and unpaired images, and the TY loss can constrain the model to generate more realistic images; 2) the discriminator $D_p$ can distinguish image pairs and make the model generate realistic deraining images. Besides, more useful information is provided by $D_p$ to enable the GAN for the adversarial training.

**Evaluation on SIRR-Data and Our Real200**

We investigate the effectiveness of our created real rainy image dataset Real200 for semi-supervised SID. As can be seen in Table 4, the deraining results of our model on Rain100L&Real200, which is a mixed dataset with Rain100L and Real200, is clearly better than that on Rain100L&SIRR-Data that is a mixed dataset with Rain100L and SIRR-Data. That is, our created new Real200 dataset can clearly enhance the deraining performance for SID by providing more rich, useful and accurate rain streak information to improve the model generalization power during the training process.

## 5 CONCLUSION

We proposed a new semi-supervised GAN-based deraining network that can use both the synthetic and real rainy images to a uniform network with two supervised and unsupervised processes. Based on the semi-supervised rain mask learner that makes the real images contribute more rain streak information to improve the results and paired discriminator, our network can clearly perform better than the recently proposed state-of-the-art method SIRR, especially on the real-world images. We also contribute a new real image dataset Real200 as a byproduct, and based on Real200 we can obtain the enhanced SID results on both synthetic and real images. Although enhanced results are obtained, we still cannot balance the generators $G_s$ and $G_r$ appropriately, since the differences of the rain streaks in synthetic and real image domains make GAN-based model difficult to converge. In the future, we will also study better training architectures and skills for cross-domain semi-supervised SID, and discover more connections between different domains.

**Table 4:** Deraining performance comparison of Semi-DerainGAN on Rain100L&SIRR-Data and Rain100L&Real200, respectively. It is clear that the results on Rain100L&Real200 are better than that on Rain100L& SIRR-Data. That is, our contributed real rain image dataset Real200 can effectively enhance the semi-supervised deraining. **Bold** denotes the best performance.

| Datasets | Rain100L& SIRR-Data | | Rain100L&Real200 | |
|---|---|---|---|---|
| Metrics | PSNR | SSIM | PSNR | SSIM |
| Our model | 34.12 | 0.958 | **35.27** | **0.988** |

## ACKNOWLEDGMENTS

This work is partially supported by the National Natural Science Foundation of China (61672365, 62072151), Anhui Provincial Natural Science Fund for Distinguished Young Scholars (2008085J30), and the Fundamental Research Funds for Central Universities of China (JZ2019HGPA0102). Both Zhao Zhang and Mingliang Xu are the corresponding authors of this paper.